\documentclass[letterpaper, 10 pt, conference]{ieeeconf}  % Comment this line out if you need a4paper

\IEEEoverridecommandlockouts                              % This command is only needed if 
\usepackage{verbatimbox}
\usepackage{color}
\usepackage{footnote}
\usepackage{bbding}
\usepackage{algorithmicx,algorithm}
\usepackage{algpseudocode}
\usepackage{graphicx}
\usepackage{subfigure}
\usepackage{amsfonts}
\usepackage{booktabs}
\usepackage{threeparttable}
\usepackage{babel}
\usepackage{hyperref}
\usepackage{amsmath}
\usepackage{booktabs} % 用于三线表格 
\usepackage{multirow} % 用于跨行单元格
\usepackage{graphicx}  % 用于resizebox命令
\usepackage{multirow}  % 用于多行单元格
\usepackage{ragged2e}
\usepackage{booktabs}
\usepackage{caption}
\usepackage{xcolor}      % 颜色支持
\usepackage{tabularx}
\usepackage{url}
\hypersetup{
    colorlinks=true,
    linkcolor=blue,
    filecolor=blue,      
    urlcolor=blue,
    citecolor=cyan,
}

\usepackage{graphics} % for pdf, bitmapped graphics files
\usepackage{graphicx}
\usepackage{float}
\title{\LARGE \bf
iGaussian: Real-Time Camera Pose Estimation via Feed-Forward 3D Gaussian Splatting Inversion  
}

\author{
 Hao Wang*, Linqing Zhao*, Xiuwei Xu, Jiwen Lu,  Haibin Yan$^\dagger$
\thanks{*Equal contribution. $^\dagger$Corresponding author.}
\thanks{
Hao Wang and Haibin Yan are with the School of Intelligent Engineering and Automation, Beijing University of Posts and Telecommunications, Beijing 100876, China. Email: \{2673439694, eyanhaibin\}@bupt.edu.cn.}
\thanks{
Linqing Zhao, Xiuwei Xu, and Jiwen Lu are with the Department of Automation, Tsinghua University, Beijing, 100084, China. Email: {zhaolinqing@mail.tsinghua.edu.cn}; {xxw21@mails.tsinghua.edu.cn}; {lujiwen@tsinghua.edu.cn}.}
}

\begin{document}

\maketitle
\thispagestyle{empty}
\pagestyle{empty}

%%%%%%%%%%%%%%%%%%%%%%%%%%%%%%%%%%%%%%%%%%%%%%%%%%%%%%%%%%%%%%%%%%%%%%%%%%%%%%%%
\begin{abstract}
Recent trends in SLAM and visual navigation have embraced 3D Gaussians as the preferred scene representation, highlighting the importance of estimating camera poses from a single image using a pre-built Gaussian model. However, existing approaches typically rely on an iterative \textit{render-compare-refine} loop, where candidate views are first rendered using NeRF or Gaussian Splatting, then compared against the target image, and finally, discrepancies are used to update the pose. This multi-round process incurs significant computational overhead, hindering real-time performance in robotics. In this paper, we propose iGaussian, a two-stage feed-forward framework that achieves real-time camera pose estimation through direct 3D Gaussian inversion. Our method first regresses a coarse 6DoF pose using a Gaussian Scene Prior-based Pose Regression Network with spatial uniform sampling and guided attention mechanisms, then refines it through feature matching and multi-model fusion. The key contribution lies in our cross-correlation module that aligns image embeddings with 3D Gaussian attributes without differentiable rendering, coupled with a Weighted Multiview Predictor that fuses features from Multiple strategically sampled viewpoints. Experimental results on the NeRF Synthetic, Mip-NeRF 360, and T\&T+DB datasets demonstrate a significant performance improvement over previous methods, reducing median rotation errors to 0.2° while achieving 2.87 FPS tracking on mobile robots, which is an impressive 10× speedup compared to optimization-based approaches. Project
 page: \url{https://github.com/pythongod-exe/iGaussian}
\end{abstract}
%%%%%%%%%%%%%%%%%%%%%%%%%%%%%%%%%%%%%%%%%%%%%%%%%%%%%%%%%%%%%%%%%%%%%%%%%%%%%%%%

%%%%%%%%%%%%%%%%%%%%%%%%%%%%%%%%%%%%%%%%%%%%%%%%%%%%%%%%%%%%%%%%%%%%%%%%%%%%%%%%
\begin{figure}[t]
    \centering
    \subfigure[{Existing Nerf-based camera pose estimation framework \cite{yen2021inerf}}]{
    	\includegraphics[width=3.10in]{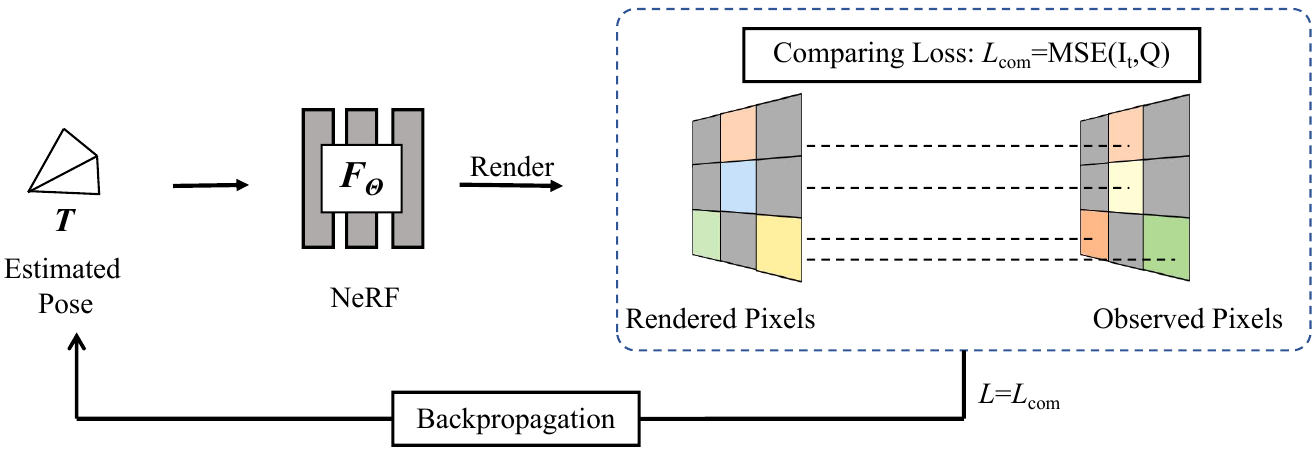}
    }\\
    \subfigure[{Existing Gaussian-based camera pose estimation framework \cite{sun2023icomma}}]{
    	\includegraphics[width=3.10in]{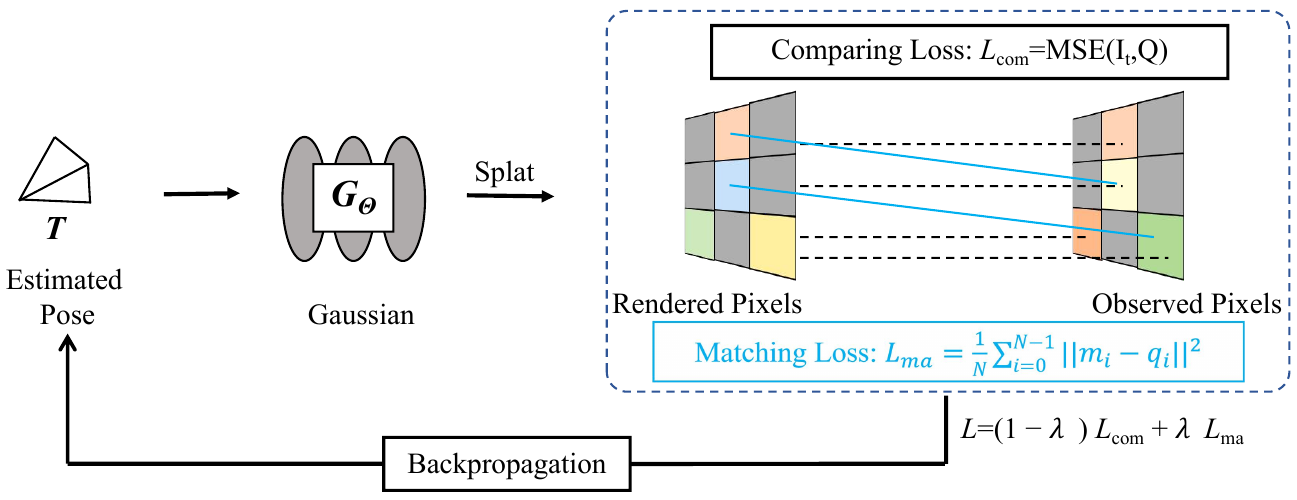}
    }\\
    \subfigure[{Our Feed-Forward framework}]{
    	\includegraphics[width=3.10in]{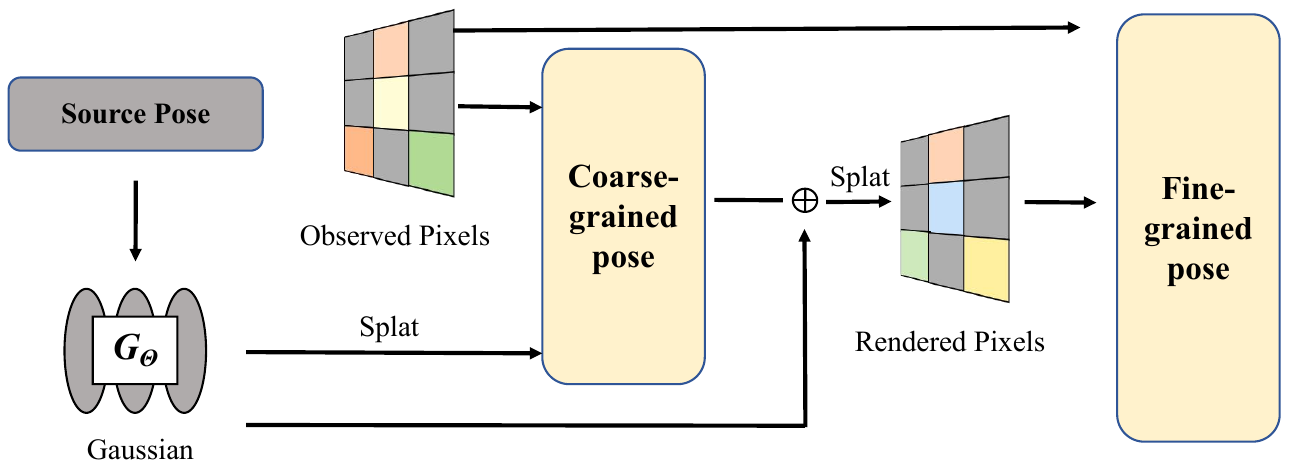}
    }
    \vspace{-3mm}
    \caption{Comparison of existing pose estimation methods based on (a) NeRF \cite{mildenhall2021nerf}, (b) 3DGS \cite{kerbl20233d}, and (c) our method.  Both (a) and (b) rely on multiple "render-compare-refine" iterations for optimization, whereas our approach follows a feed-forward paradigm.\justifying}
\vspace{-6mm}
\label{fig:1}
\end{figure}
%%%%%%%%%%%%%%%%%%%%%%%%%%%%%%%%%%%%%%%%%%%%%%%%%%%%%%%%%%%%%%%%%%%%%%%%%%%%%%%%

%%%%%%%%%%%%%%%%%%%%%%%%%%%%%%%%%%%%%%%%%%%%%%%%%%%%%%%%%%%%%%%%%%%%%%%%%%%%%%%%
\section{Introduction}
Estimating camera poses from a single image using a pre-built 3D scene representation is a critical capability for applications such as robotic navigation and augmented reality (AR), where real-time localization in unknown environments is paramount. This task involves inferring the 6-DoF camera pose (rotation and translation) of an observed image relative to a pre-optimized 3D Gaussian splatting model, which is a lightweight, differentiable representation that enables efficient scene rendering. Such a capability eliminates the need for repeated scene reconstruction during deployment, allowing robots or AR devices to localize instantly in environments where prior mapping has been completed. 

The fundamental challenge lies in bridging the gap between the expressive power of 3D Gaussian representations and the need for feed-forward pose inference. Existing methods typically rely on iterative optimization frameworks that render candidate views, compare them against the target image, and refine poses through gradient-based updates, a computationally expensive process incompatible with real-time requirements. While Gaussians enable photorealistic rendering, their unstructured nature complicates direct geometric reasoning, and iterative refinement introduces latency that scales poorly with scene complexity. 

While Neural Radiance Fields (NeRF) \cite{mildenhall2021nerf} and subsequent works \cite{liu2020neural} achieve accurate scene reconstruction via photometric optimization, their reliance on volumetric rendering and iterative pose refinement results in impractical computational costs. Recent 3D Gaussian-based SLAM systems \cite{yu2021pixelnerf} address rendering efficiency through splatting but inherit two core limitations: (1) Persistent dependency on slow "render-compare-refine" loops for joint Gaussian-camera optimization, and (2) Requirement for depth sensors to bootstrap geometry, restricting deployment to RGB-only platforms. Optimization-driven methods like iNeRF \cite{yen2021inerf} and iComMa \cite{sun2023icomma} further exemplify this trade-off—while achieving sub-degree pose accuracy, they demand hundreds of iterations per frame, rendering them unsuitable for real-time robotics. These constraints highlight an unresolved tension between accuracy and efficiency in pose estimation.

In this paper, we aim to bridge through a feed-forward inversion paradigm that directly maps image features to camera poses, bypassing both iterative optimization and depth dependency. We first introduce a spherical sampling strategy to enable efficient camera pose estimation from pre-trained 3D Gaussian scene models. At its core, the method uniformly samples multiple candidate viewpoints on a sphere encompassing the target scene, generating synthetic reference images through Gaussian splatting. These reference images, paired with the observed target image, are fed into a pose regression network that directly predicts a coarse 6DoF camera pose in a single forward pass. This network leverages cross-view geometric relationships through an attention mechanism, focusing on spatially consistent features across sampled viewpoints to resolve pose ambiguities. To refine the initial prediction, we employ a hybrid optimization stage: A synthetic image is rendered using the coarse pose guides feature matching with the target image, while a vision transformer resolves residual pose errors by analyzing relative spatial transformations. Crucially, the entire pipeline operates without iterative optimization or depth sensors. Experiments show that our method outperforms optimization-driven baselines both in terms of performance and speed.
%%%%%%%%%%%%%%%%%%%%%%%%%%%%%%%%%%%%%%%%%%%%%%%%%%%%%%%%%%%%%%%%%%%%%%%%%%%%%%%%
\begin{figure*}[ht]
    \centering
    \includegraphics[scale=0.52]{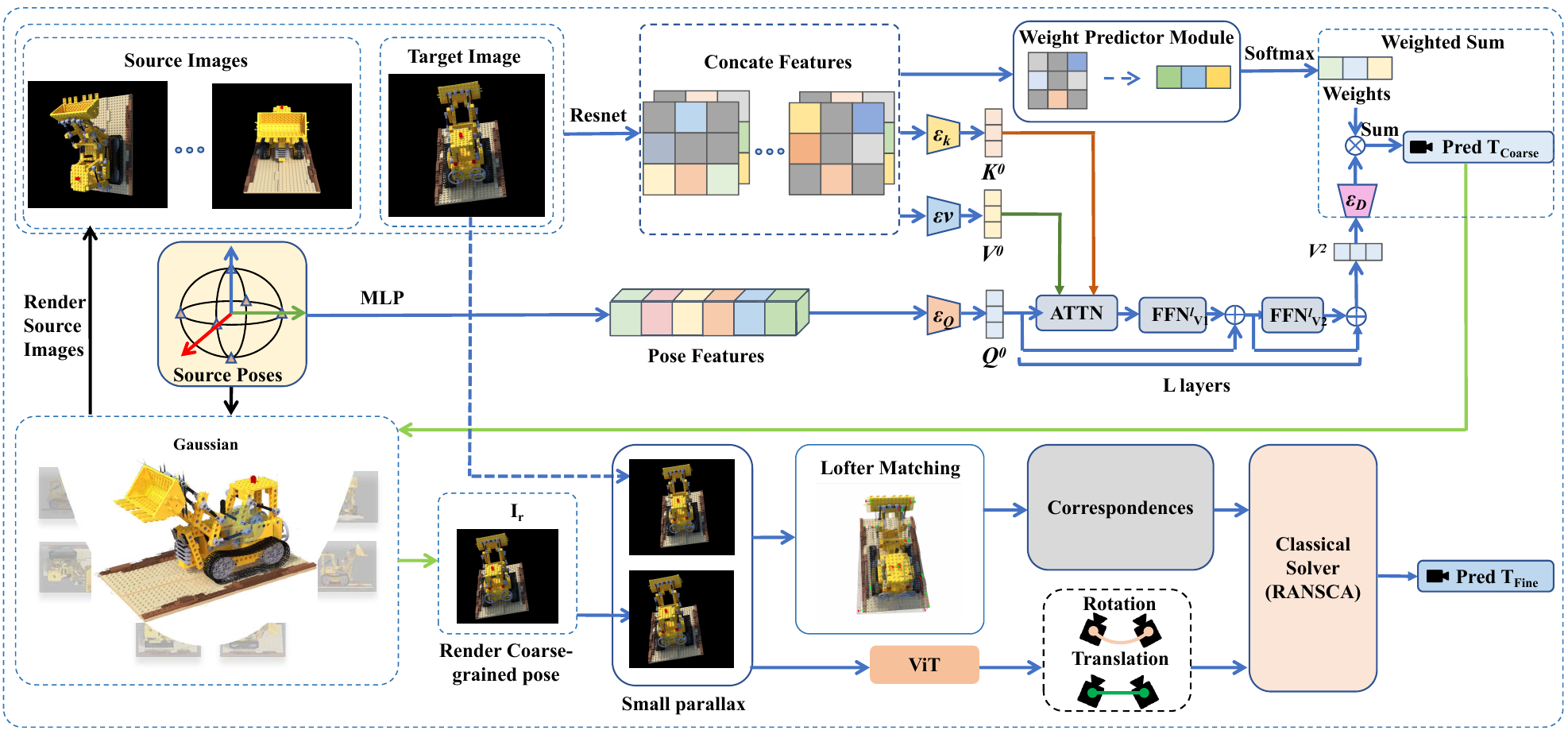}
    \caption{Overview of iGaussian pipeline. Our approach estimates the camera pose $T_{fine}$  of an observed image \( I \) using a two-stage pipeline. First, a Pose Attention Network predicts a coarse 6DoF pose \( (R, t) \) from the target image and generates a reference view using a 3D Gaussian representation. Then, a matching and solver module refines the pose by computing the relative transformation between the observed and rendered images. A Transformer-based ViT predicts translation scale and learned correspondence estimation with geometric constraints to enhance accuracy. The framework eliminates iterative rendering, ensuring efficient and precise pose regression.\justifying}
    \label{fig:pipeline}
    \vspace{-0.4cm}
\end{figure*}
%%%

\section{Related Work}
\noindent \textbf{Learning-based Camera Pose Estimation.} Recent advances in deep learning have revolutionized camera pose estimation by replacing traditional geometric pipelines with data-driven frameworks. While early attempts concatenated multi-view image features for direct pose regression \cite{kendall2015posenet,balntas2018relocnet,brahmbhatt2018geometry}, these suffered from limited generalization and scale ambiguity. Subsequent advances introduced correlation volumes \cite{teed2021droid} and transformer architectures \cite{dosovitskiy2020image,rockwell20228} to enable more sophisticated cross-view feature interaction. Recent methods integrate geometric priors into neural networks through architectural innovations - 8-Point ViT \cite{rockwell20228} embeds fundamental matrix constraints into vision transformers, while RPNet \cite{en2018rpnet} combines absolute pose regression with geometric verification. Parallel efforts focus on reconstructing planar 3D structures from minimal wide-baseline image pairs \cite{tan2023nope,qian2020associative3d}, demonstrating the power of geometric-aware learning. Our approach draws inspiration from FAR \cite{rockwell2024far}, which establishes a hybrid framework combining learned correspondence priors with robust solver optimization. Feature matching-based pose estimation remains a fundamental problem in computer vision, with classical approaches relying on handcrafted descriptors (e.g., SIFT \cite{lowe2004distinctive}) combined with RANSAC \cite{sarlin2021back} and the 8-point algorithm \cite{longuet1981computer,nister2004efficient}. While these methods demonstrate robustness to noise, they often struggle in views with extreme viewpoint variations or low-texture surfaces. Recent advancements have introduced learning-based frameworks to overcome these limitations. Methods such as SuperGlue \cite{sarlin2020superglue}, LoFTR \cite{sun2021loftr}, and MatchFormer \cite{wang2022matchformer} leverage deep neural networks to enhance feature correspondence quality, improving robustness across challenging conditions. Additionally, approaches like DeepIM \cite{li2018deepim} and MOPED \cite{kolker2012moped} have enabled real-time 6D pose estimation by integrating learning-based refinement with geometric reasoning. Beyond image-based matching, several methods focus on aligning images with target point clouds or 3D models \cite{he2022onepose++,fan2023pope}, achieving good performance in scenarios requiring accurate spatial alignment.

\noindent \textbf{NeRF and Gaussian Splatting for Pose Estimation.} Implicit neural representations like NeRF and 3DGS have revolutionized simultaneous localization and mapping (SLAM) by unifying scene reconstruction and camera tracking within a differentiable framework. NeRF-based SLAM systems \cite{zhu2022nice,johari2023eslam,rosinol2023nerf} and iNeRF \cite{yen2021inerf} jointly optimize camera poses with radiance fields through photometric consistency in volume rendering. In contrast, 3DGS \cite{kerbl20233d} parameterizes scenes as anisotropic 3D Gaussians with explicit spatial control, enabling efficient rasterization via splatting-based rendering and dynamic scene adaptation through Gaussian attribute optimization. Recent advancements like Gaussian Splatting SLAM \cite{matsuki2024gaussian} and SemGauss-SLAM \cite{zhu2024semgauss} achieve real-time free-viewpoint streaming by training Gaussians on the fly, reducing computational latency by an order of magnitude compared to NeRF-based systems. These approaches demonstrate enhanced scalability and geometric fidelity over traditional discrete representations \cite{tremblay2018deep}, particularly in preserving high-frequency details and supporting dynamic object handling through techniques. However, challenges persist in textureless region reconstruction and extreme occlusion scenarios due to the dependency on photometric cues. Our framework in Fig.~\ref{fig:1} addresses these limitations, and implements a coarse-to-fine Gaussian selection strategy to prioritize geometrically reliable regions during optimization. Our methodology similarly adopts 3DGS as a scene prior to guide pose estimation, akin to iComMa \cite{sun2023icomma}, leveraging its powerful rendering capabilities to achieve robust online operational performance.
%%%%%%%%%%%%%%%%%%%%%%%%%%%%%%%%%%%%%%%%%%%%%%%%%%%%%%%%%%%%%%%%%%%%%%%%%%%%%%%%

%%%%%%%%%%%%%%%%%%%%%%%%%%%%%%%%%%%%%%%%%%%%%%%%%%%%%%%%%%%%%%%%%%%%%%%%%%%%%%%%

\section{Approach}
We propose iGuassian, which aims to predict the camera pose of the observed viewpoint from a 3D Gaussian scene representation. Assuming that the 3D Gaussian \cite{kerbl20233d} model of the scene or object is represented by parameters $\Theta$, and the camera's intrinsic parameters are known, but the camera pose $T$ of the target image $I$ remains unknown. Our goal is to solve for the camera pose $T_{fine}$ of the observed image $I$. Our method can infer the camera pose $T_{fine}$ directly from the input image and scene parameters $\Theta$, independent of reference images.

iGaussian combines the strengths of correspondence estimation and end-to-end pose estimation, leveraging a 3D Gaussian model as a prior for precise pose estimation. As shown in Fig.~\ref{fig:pipeline}, our method operates in two stages. First, the Rendering-Based Coarse-Grained Pose Estimation network regresses a coarse 6DoF pose $(R, t)$ from the observed image $I$ and renders a reference view $I_r$ using a 3D Gaussian rendering pipeline. Second, a matching and solver module computes the relative pose $\Delta T$ between ${I}$ and $I_r$, guided by a Transformer-based ViT  \cite{rockwell20228,vaswani2017attention} that predicts the $\text{Sacle}$. This architecture eliminates the need for iterative ``compare-render-compare'' steps, relying instead on a single rendering and high-quality correspondences for efficient and accurate pose estimation.

\subsection{Rendering-Based Coarse-Grained Pose Estimation}
\label{subsec:A}
Given a target image ${I}$ and a 3D Gaussian representation $\Theta$, our network estimates the absolute pose by leveraging spatial constraints and source image pose information to map input to predicted pose. Unlike traditional view synthesis networks, our approach focuses on learning 3D positional relationships rather than pixel-level transformations, enabling precise 6DoF pose prediction and effective cross-viewpoint pose distribution modeling for accurate camera pose regression.

\begin{figure}[t]
    \centering
    \subfigure[Object-level sampling strategy]{
    	\includegraphics[width=3.10in]{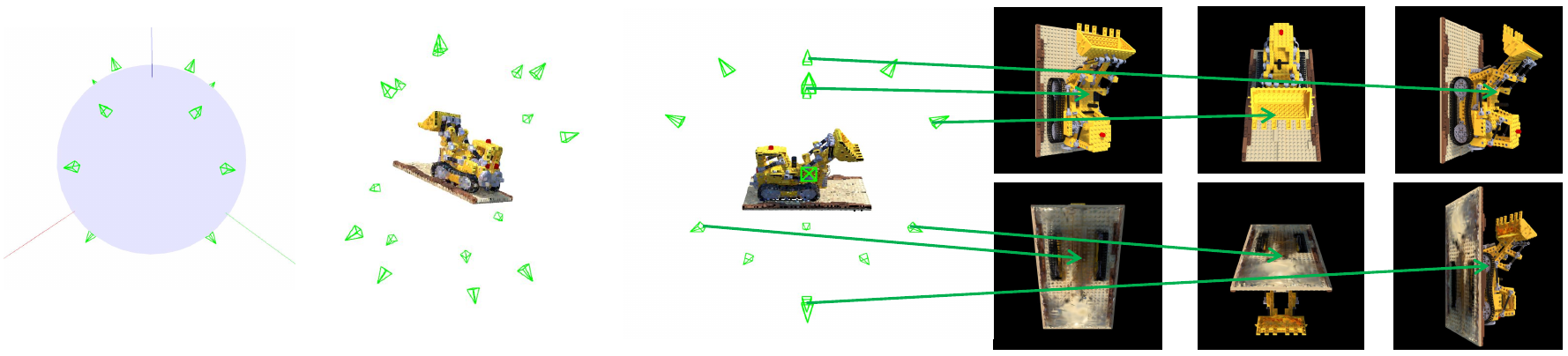}
    }\\
    \subfigure[Scene-level sampling strategy]{
    	\includegraphics[width=3.10in]{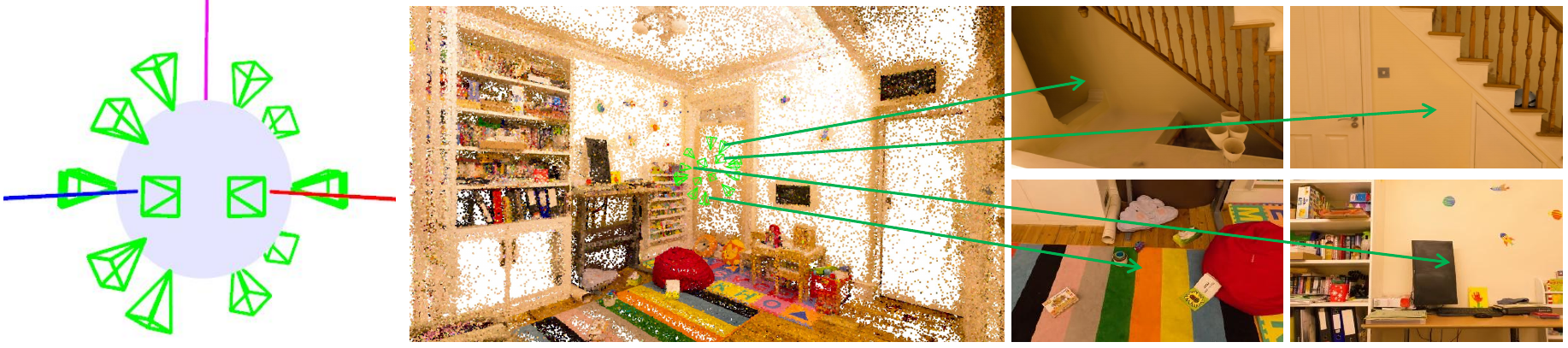}
    }
    \vspace{-3mm}  
    \caption{Object-level and scene-level sampling strategies. (a) and (b) illustrate different sampling strategies for objects and scenes, respectively. In both cases, cameras are uniformly distributed on a spherical surface. However, for objects, the camera viewpoints are always directed toward the object’s center, ensuring comprehensive coverage. In contrast, for scenes, the camera viewpoints consistently face away from the scene’s center, capturing a broader environmental context.\justifying}
    \label{fig:3}
    \vspace{-4mm}    
\end{figure}

\paragraph{Spatial Uniform Sampling Strategy} Since our input merely encompasses the target image ${I}$ and the 3D Gaussian representation $\Theta$ of an object or scene, we developed a spatial uniform sampling strategy to sample source viewpoint cameras from the Gaussian model as a reference. To ensure that the sampled source viewpoints uniformly cover the entire 3D scene, we constructed a viewpoint distribution strategy based on the 3D Gaussian model and spherical coordinate system in Fig.~\ref{fig:3}.
Specifically, we define a spherical trajectory centered at the coordinate origin with a fixed radius $R$, allowing cameras to be uniformly distributed on the sphere. The radius R is adaptively adjusted based on different scenes to ensure optimal viewing distances and coverage for various object scales and scene complexities. By adjusting the polar angle $\mathrm{\Phi}$ and azimuthal angle $\psi$, we achieve uniform sampling. In the spherical coordinate system, the 3D position $p=(x,y,z)$ of the camera is calculated as:
\begin{equation}
\begin{cases}
x=R\cdotp\sin\left(\Phi\right)\cdotp\cos\left(\psi\right)\\
y=R\cdotp\sin\left(\Phi\right)\cdotp\sin\left(\psi\right)\\
z=R\cdotp\cos\left(\Phi\right),
\end{cases}
\end{equation}
where $R$ is the spherical radius, $\theta$ is the polar angle, and $\psi$ is the azimuthal angle. To ensure each sampled camera viewpoint points toward the object's center (the origin), we generate the view matrix $\mathbf{M}$ using the $look\_at$ function \cite{foley1996computer}:
\begin{equation}
\mathbf{M} = look\_at\left(p,\, target,\, up\right),
\end{equation}
where $p$ is the camera position, $target$ is the focal point, and $up=(0,1,0)$ or $(0,-1,0)$ defines the upward direction. By controlling the sampling points of $\theta$ and $\psi$, we generate camera positions and compute their corresponding view matrices. These matrices are used to render source images $\{{x}_s^{i}\}_{i=1}^N$, forming a complete source viewpoint set $\mathcal{S}$. The target image ${I}$, along with the $N$ source images, is input into the network for pose regression:
\begin{equation}
T = f_\theta\left({I}, \{{x}_s^{i}\}_{i=1}^N; \theta\right).
\end{equation}
This approach ensures that the camera trajectory is independent of scene complexity, allowing the same strategy to be applied to scenes of varying scales and complexities without requiring additional depth information estimation.

\paragraph{Attention-based Spatial Transformation Learning} In the absence of appropriate inductive biases, networks trained solely on images may struggle to generalize, particularly in tasks involving direct learning of geometric transformations rather than pixel-level mapping. Directly regressing pose from image features without explicit spatial correspondences often leads to unstable or imprecise predictions. To address this, we propose a method that guides self-attention for spatial transformation learning, departing from traditional appearance-based approaches \cite{ren2022adela}. Specifically, we encode the pose of source images as query vectors $Q^i$ and direct it to focus on the most critical components for pose estimation within the concatenated features $K^i$ and $V^i$ formally:
% 更规范的数学表达式写法
\begin{flalign}
&\mathrm{K}^{i},\ \mathrm{V}^{i} = \text{concat}(\phi_r(I), \phi_r(x_s^i)), \\
&\mathrm{Q}^{i} = \text{MLP}(\mathrm{Pose}(x_s^i)), \\
&\mathrm{Attention}(\mathrm{Q}^i,\ \mathrm{K}^i,\ \mathrm{V}^i) = \text{softmax}\left( \frac{\mathrm{Q}^i (\mathrm{K}^i)^T}{\sqrt{d_k}} \right) \mathrm{V}^i,
\end{flalign}
where $\phi_r$ is the image feature extractor (a pre-trained ResNet in this work), and ${Pose}(x_s^i)$ represents the pose information of source images. The query vector ${Q}^i$ is projected into a 128-dimensional embedding via an MLP. The attention output is refined through feed-forward networks (FFNs) and decoded to the image space:
\begin{flalign}
& V^{i_1} = \text{FFN}_1 (\text{Attention}(Q^i, K^i, V^i)) + V^i, \\
& V^{i_2} = \text{FFN}_2 (V^{i_1}), \\
& \hat{T}^i = \xi_D (V^{i_2}).
\end{flalign}
\label{eq: ProjHighDim}
By concatenating target and source image features as $K^i$ and $V^i$, we enhance cross-view geometric information, improving spatial alignment and providing stable geometric cues. The MLP-projected query vectors $Q^i$ explicitly guide attention to regions relevant to target viewpoint changes, ensuring stable attention distribution and mitigating feature misalignment caused by viewpoint variations.

 \paragraph{Multi-viewpoint Feature Fusion and Pose Regression} Given a set of source images $\{x_s^i\}_{i=1}^N$ and a target image $I$, we first extract and concatenate their features and then predict multiple candidate poses using a pose regression network:
\begin{equation}
\widehat{T}^i=f(x_s^i,\mathrm{I;~}\theta),
\end{equation}
where $x_s^i$ and $I$ represent the source and target images, respectively, and $f(.)$ is the pose regression network predicting the camera pose $\widehat{T}^i$. To address the imbalance in information provided by different source viewpoints, we designed the Weight Predictor Module (WPM), which assigns weights to each predicted pose based on feature information. The weights $w^i$ are predicted using an MLP:
\begin{equation}
w^i=g(f(x_s^i,I)),
\label{eq: WPM}
\end{equation}
where $g(.)$ is the weight prediction network. The weights are normalized via Softmax to form a probability distribution:
\begin{equation}
w^i = \frac{\exp(w^i)}{\sum_{j=1}^{N} \exp(w^j)} g(f(x_s^i, I)),
\label{eq: GenProWeight}
\end{equation}
The normalized weights $w^i$ indicate the importance of each $x_s^i$ for candidate pose prediction. The final predicted pose $T_{coarse}$ is computed as a weighted average:
\begin{equation}
T_{coarse} = \sum_{i=1}^{N} w^i \cdot \hat{T}^i.
\label{eq: CalFinalPose}
\end{equation}
Through spherical uniform sampling, we ensure comprehensive coverage of source viewpoints, reducing ambiguity in pose estimation. The WPM enhances prediction accuracy by assigning higher weights to more reliable viewpoints.

\subsection{Correspondence-Based Pose Optimization}
\label{subsec:A}

After predicting an initial pose $T_{coarse}$ using a Gaussian scene prior-based pose regression network and rendering an image $I_r$ via the 3D Gaussian rendering pipeline, we obtain a rendered image close to the target image $I$. Although the visual difference between $I_r$ and $I$ is small, further refinement is necessary. In our method, relative pose optimization is a critical step to enhance the accuracy of the initial pose prediction $T_{coarse}$, aligning $I$ and $I_r$ more precisely. This process integrates the translation scale predicted by ViT with traditional feature matching (LoFTR \cite{sun2021loftr} and RANSAC \cite{fischler1981random}), achieving accurate optimization of the predicted pose $T_{coarse}$.

\paragraph{Feature Point Matching and Relative Pose Calculation.} In correspondence-based pose estimation, methods like RANSAC and its variants are commonly used. These methods randomly sample minimal point sets and fit models using n-point algorithms, with Sampson error as the inlier threshold. Given a set of 2D correspondences $M{=}\{(p,q)\}$, the scoring function for hypothesis $H$ \cite{torr2000mlesac} is:
\begin{equation} 
{score}(H) = \sum_{(p,q) \in M} 1 \left( E(p, q | H) < \sigma \right),
\label{eq:ScoreFunc}
\end{equation}
where $E(p,q|H)$ is the Sampson error, the model with the highest score (most inliers) is selected. This sampling process repeats $N$ times until convergence.
Under the initial pose prediction $T_{coarse}$, feature point matching computes the corresponding feature set $\{(p_i,q_i)\}$ between the source image $I_r$ and target image $I$ where $p_i$ and $q_i$ are matching points. The close viewpoints yield abundant matches, enabling precise relative pose estimation, with RANSAC solving for the relative pose $\Delta T$:
\begin{equation} 
\Delta T=\text{RANSAC}({p_i,q_i}).
\label{eq:RANSAC}
\end{equation}
The resulting relative pose $\Delta T$ includes the rotation matrix $\Delta R$ and translation vector $\Delta t$, but lacks translation scale. Traditional geometric methods like RANSAC and the 8-Point Algorithm excel in small-disparity views, where high viewpoint overlap ensures accurate results. In our approach, the Gaussian scene prior-based pose regression network provides an accurate initial pose $T_{coarse}$ and rendered image $I_r$, enabling RANSAC to compute with high precision.

% \subsubsection{Translation Scale Correction}
\paragraph{Translation Scale Correction}
Since the relative pose $\Delta T$ obtained via RANSAC lacks translation scale, we employ ViT which is trained with absolute pose to predict the relative pose $T_{vit}$ with translation scale between the target and rendered images. This enhances the solver's robustness by integrating learning-based predictions. The translation scale $|t_{vit}|$ is used to correct $\Delta T$:
\begin{equation} 
|t_{fine}|=|\Delta t|\times|t_{vit}|.
\label{eq:RecT}
\end{equation}
The corrected $T_{fine}$ provides a more accurate representation of the relative pose between the source and target images. Traditional geometric methods excel in small-disparity views but struggle with translation scale and complex scenes. Deep learning methods like ViT adapt better to complex scenes but rely on training data and architecture. By combining their strengths, our approach achieves multi-stage pose optimization, enhancing accuracy and stability. 

\subsection{Losses}
\label{subsec:A}
We define a pose loss function that combines rotation and translation errors.
\paragraph{Rotation Error} The rotation error is computed as the angular difference between the predicted and ground truth rotation matrices. Given two unit quaternions $q_{fine}$ and $q$ representing the predicted and ground truth rotations, respectively, the rotation error ${L}_{rot}$ is defined as:
\begin{equation}
{L}_{rot} = 2 \cdot \arccos(|q_{fine} \cdot q|) \cdot \frac{180}{\pi},
\end{equation}
where $|q_{fine} \cdot q|$ denotes the dot product of the two quaternions.
\paragraph{Translation Error}
The translation error ${L}_{trans}$ measures the Euclidean distance between the predicted translation vector $t_{fine}$ and the ground truth translation vector $t$:
\begin{equation}
L_{trans}=\|t_{fine}-t\|_2,
\end{equation}
where $\left\|\cdot\right\|_{2}$ denotes the $L_{2-\mathrm{norm}}$.
\paragraph{Overall Loss Function} 
The overall pose loss ${L}_{pose}$ is a weighted combination of the rotation and translation errors: 
\begin{equation}
{L}_{pose}=\lambda_r\cdot{L}_{{rot}}+\lambda_t\cdot{L}_{{trans}},
\end{equation}
where $\lambda_{r}$ and $\lambda_{t}$ are hyperparameters that control the relative importance of the rotation and translation errors, respectively. In our implementation, we set $\lambda_{r}=1.0$, $\lambda_{t}=30$.

%%%%%%%%%%%%%%%%%%%%%%%%%%%%%%%%%%%%%%%%%%%%%%%%%%%%%%%%%%%%%%%%%%%%%%%%%%%%%%%%

%%%%%%%%%%%%%%%%%%%%%%%%%%%%%%%%%%%%%%%%%%%%%%%%%%%%%%%%%%%%%%%%%%%%%%%%%%%%%%%% table1 setting
\captionsetup[table]{skip=2pt, singlelinecheck=false, justification=raggedright, labelfont=bf, labelsep=period}
%%%%%%%%%%%%%%%%%%%%%%%%%%%%%%%%%%%%%%%%%%%%%%%%%%%%%%%%%%%%%%%%%%%%%%%%%%%%%%%% table1
\begin{table*}[t]
    \centering
    \footnotesize
    \setlength{\tabcolsep}{3mm}
    \caption{\textbf{Comparison of Accuracy and Runtime on the NeRF Synthetic Dataset.} Our method maintains superior success rates and the fastest inference across angular ranges, demonstrating strong robustness in low-overlap views.\justifying}
    \label{tab:Synthetic}
    \begin{tabular}{cccc|ccc|ccc}
        \toprule
        \multirow{2}{*}{\textbf{Methods}} & \multicolumn{3}{c}{±[20°, 40°]} & \multicolumn{3}{c}{±[40°, 80°]} & \multicolumn{3}{c}{±[80°, 180°]}\\
        \cmidrule(lr){2-4} \cmidrule(lr){5-7} \cmidrule(lr){8-10}
        & iNeRF & iComMa & Ours & iNeRF & iComMa & Ours & iNeRF & iComMa & Ours \\
        \midrule
        Success Rate (\%) & 62.94 & 78.69 & \textbf{82.06} & 17.5  & 45.44 & \textbf{82.31} & 0  & 0 & \textbf{81.86}\\
        Time (s)         & 23.33 & 3.95  & \textbf{0.38}  & 26.7   & 10.43 & \textbf{0.38} & -- & --& \textbf{0.38} \\
        \bottomrule
    \end{tabular}
    \vspace{-5mm} 
\end{table*}
%%%%%%%%%%%%%%%%%%%%%%%%%%%%%%%%%%%%%%%%%%%%%%%%%%%%%%%%%%%%%%%%%%%%%%%%%%%%%%%% table1

%%%%%%%%%%%%%%%%%%%%%%%%%%%%%%%%%%%%%%%%%%%%%%%%%%%%%%%%%%%%%%%%%%%%%%%%%%%%%%%% table2
\begin{table}[t]
    \centering
    \caption{\textbf{Pure Rotation Test on the Mip-NeRF360 Dataset.} The error metrics for SIFT and SuperPoint are computed exclusively on successful correspondences. Their failure rates exceed 50\%.\justifying}
    \vspace{1pt}
    \label{tab:2}
    \resizebox{1.05\columnwidth}{!}{ 
    \begin{tabular}{lccccccccc} % 8列定义
        \toprule
        \multirow{3}{*}{Methods} & 
        \multicolumn{3}{c}{±[20°,40°]} & 
        \multicolumn{3}{c}{±[40°,80°]} & 
        \multicolumn{3}{c}{±[80°,180°]}\\
        \cmidrule(lr){2-4} \cmidrule(lr){5-7} \cmidrule(lr){8-10}
        & Avg.$\downarrow$ & Med.$\downarrow$ & 
        $\leq15^\circ\uparrow$ & Avg.$\downarrow$ & Med.$\downarrow$ & $\leq15^\circ\uparrow$ & Avg.$\downarrow$ & Med.$\downarrow$ & $\leq15^\circ\uparrow$\\
        \midrule
        SIFT\* & 18.9 & 3.13 & 22.4 & 38.8 & 13.8 & 5.7 & -- & -- & 0 \\ % 新增空列
        SuperPoint\* & 6.38 & 1.79 & 16.5 & 6.8 & 6.85 & 2.3 & -- & -- & 0 \\
        LoFTR & 3.36 & 0.84 & 92.3 & 28.62 & 9.24 & 57.4 & 48.2 & 42,7 & 1.5 \\
        8-Point ViT & 4.56 & 2.98 & 94.1 & 9.82 & 4.79 & 76.4 & 32.43 & 17.23 & 44.6  \\
        FAR & 1.87 & 0.34 & \textbf{100} & 4.48 & 1.16 & 94.4 & 22.61 & 14.58 & 52.8 \\
        Ours & \textbf{0.57} & \textbf{0.22} & \textbf{100} & \textbf{0.54} & \textbf{0.23} & \textbf{100} & \textbf{0.55} & \textbf{0.23} & \textbf{100} \\ 
        \bottomrule
    \end{tabular}
    }
    \vspace{-3mm}
\end{table}
%%%%%%%%%%%%%%%%%%%%%%%%%%%%%%%%%%%%%%%%%%%%%%%%%%%%%%%%%%%%%%%%%%%%%%%%%%%%%%%% table2

%%%%%%%%%%%%%%%%%%%%%%%%%%%%%%%%%%%%%%%%%%%%%%%%%%%%%%%%%%%%%%%%%%%%%%%%%%%%%%%% table3
\begin{table*}[t]
    \centering
    % \raggedright
    \caption{\textbf{Evaluation of Rotational-Translational Performance on the T\&T+DB Dataset.} End-to-end methods (ViT-8-Point) excel in translation estimation but exhibit deficiencies in rotation estimation. Solver-based methods (LoFTR) perform well in rotation estimation. Our hybrid approach strikes a balance between the two modalities and achieves improvements of over 30\% in both translation and rotation performance compared to FAR, which also combines the two modalities. \justifying}
    \vspace{2pt}
    \setlength{\tabcolsep}{0mm}
    \label{tab:3}
    \resizebox{\textwidth}{!}{%
    \begin{tabular}{lccccccccccccccccccc}
        \toprule
        \multirow{3}{*}{Methods} & 
        \multicolumn{6}{c}{$\pm[20^{\circ}, 40^{\circ}]$} & 
        \multicolumn{6}{c}{$\pm[40^{\circ}, 80^{\circ}]$} & 
        \multicolumn{6}{c}{$\pm[80^{\circ}, 180^{\circ}]$} \\
        \cmidrule(lr){2-7} \cmidrule(lr){8-13} \cmidrule(lr){14-19}
        & \multicolumn{3}{c}{Translation (m)} & \multicolumn{3}{c}{Rotation ($^{\circ}$)} & \multicolumn{3}{c}{Translation (m)} & \multicolumn{3}{c}{Rotation ($^{\circ}$)} & \multicolumn{3}{c}{Translation (m)} & \multicolumn{3}{c}{Rotation ($^{\circ}$)} \\
        \cmidrule(lr){2-4} \cmidrule(lr){5-7} \cmidrule(lr){8-10} \cmidrule(lr){11-13} \cmidrule(lr){14-16} \cmidrule(lr){16-19} 
        & Avg. $\downarrow$ & Med. $\downarrow$ & $\leq$0.5m $\uparrow$ & Avg. $\downarrow$ & Med. $\downarrow$ & $\leq$15$^{\circ}$ $\uparrow$ & Avg. $\downarrow$ & Med. $\downarrow$ & $\leq$0.5m $\uparrow$ & Avg. $\downarrow$ & Med. $\downarrow$ & $\leq$15$^{\circ}$ $\uparrow$ & Avg. $\downarrow$ & Med. $\downarrow$ & $\leq$0.5m $\uparrow$ & Avg. $\downarrow$ & Med. $\downarrow$ & $\leq$15$^{\circ}$ $\uparrow$\\
        \midrule
        SuperGlue & -- & -- & -- & 3.88 & 0.59 & 97 & -- & -- & -- & 17.4 & 9.33 & 68 & -- & -- & 0 & -- & -- & 0 \\
        NOPE-SAC-Reg & 0.34 & 0.26 & 90.4 & 2.77 & 0.82 & 98.2 & 0.47 & 0.35 & 75.8 & 7.53 & 2.49 & 92.4 & 1.15 & 0.94 & 7.2 & 34.6 & 29.5 & 14.1\\
        LoFTR+Reg.Scale & 0.44 & 0.28 & 73.2 & 2.92 & 0.64 & 96.8 & 0.78 & 0.52 & 46.3 & 13.9 & 5.14 & 80.1 & 1.41 & 1.35 & 2.5 & 49.2 & 46,5 & 1.3\\
        8-Point ViT & 0.38 & 0.28 & 87.3 & 3.42 & 1.83 & 97.9 & 0.52 & 0.33 & 72.3 & 8.23 & 3.85 & 86.6 & 0.92 & 0.84 & 12.3 & 28.23 & 23.85 & 26.6\\
        FAR & 0.24 & 0.15 & \textbf{100} & 1.23 & 0.38 & \textbf{100} & 0.38 & 0.21 & \textbf{100} & 3.07 & 0.86 & \textbf{100} & 0.74 & 0.57 & 45.9 & 20.1 & 14.6 & 52.7 \\
        \textbf{Ours} & \textbf{0.11} & \textbf{0.08} & \textbf{100} & \textbf{0.43} & \textbf{0.2} & \textbf{100} & \textbf{0.09} & \textbf{0.07} & \textbf{100} & \textbf{0.45} & \textbf{0.2} & \textbf{100} & \textbf{0.09} & \textbf{0.07} & \textbf{100} & \textbf{0.45} & \textbf{0.2} & \textbf{100}\\
        \bottomrule
    \end{tabular} 
    }
    \vspace{-8pt}
\end{table*}
%%%%%%%%%%%%%%%%%%%%%%%%%%%%%%%%%%%%%%%%%%%%%%%%%%%%%%%%%%%%%%%%%%%%%%%%%%%%%%%% table3

%%%%%%%%%%%%%%%%%%%%%%%%%%%%%%%%%%%%%%%%%%%%%%%%%%%%%%%%%%%%%%%%%%%%%%%%%%%%%%%%
\section{Experiment}
We conducted a series of experiments to evaluate the performance of our proposed method in the 6DoF pose estimation task and compared it with SOTA approaches. We tested it on multiple benchmark datasets. Our model is trained jointly across all datasets without distinguishing specific scenes, enabling unified training and enhanced generalization capabilities across diverse environments. First, we test it on 8 objects from NeRF’s synthetic dataset \cite{mildenhall2021nerf}, which primarily consist of single-object scenes, providing a controlled environment for performance evaluation. Additionally, we employed 360-degree unbounded scene datasets from Mip-NeRF360 \cite{barron2022mip}, which offer a more comprehensive reflection of the model's capabilities in large-scale and complex environments. To further verify the model’s generalization ability in real-world scenarios, we introduced the $\mathrm{T \& T+DB}$ \cite{hedman2018deep} dataset, which represents typical 360-degree scenes and includes various real-world environmental variations. We also conducted ablation studies to assess the contribution of different modules to the final performance. In all experiments, we uniformly used 16 source images as input, selected based on a specific sampling strategy, where the elevation angle $\mathrm{\Phi}$ was set to $\pm[30^\circ,60^\circ]$ and the azimuth angle $\psi$ was sampled at $[0^\circ,90^\circ,180^\circ,270^\circ]$. Additionally, we have employed changes in color and brightness to enhance the images.

\subsection{NeRF Synthetic Dataset}
\setcounter{paragraph}{0} % 重新从 'a' 开始编号
\label{subsec:A}
\paragraph{Setting}We evaluated iGaussian on 8 objects from the NeRF Synthetic Dataset, comparing it with INeRF \cite{yen2021inerf} and iComMa \cite{sun2023icomma}. INeRF parameters were set to batch\_size=2048 and sampling\_strategy=interest\_regions. The test datasets was generated by applying translation and rotation transformations to source poses, rendering target images using a pre-trained 3D Gaussian model trained for 30,000 iterations. For each scene, 300 target images were generated: 100 with angular errors in $\pm[20^{\circ}, 40^{\circ}]$ and 100 in $\pm[40^{\circ}, 80^{\circ}]$, and 100 in $\pm[80^{\circ}, 180^{\circ}]$ with a translation offset of $[-0.2, 0.2]$. INeRF and iComMa used target-source image pairs, while iGaussian required only the target image. while iGaussian used a Gaussian sampling strategy with $R=4$ and camera viewpoints directed toward the scene origin.

\begin{figure}[t]
    \centering
    \includegraphics[width=\columnwidth]{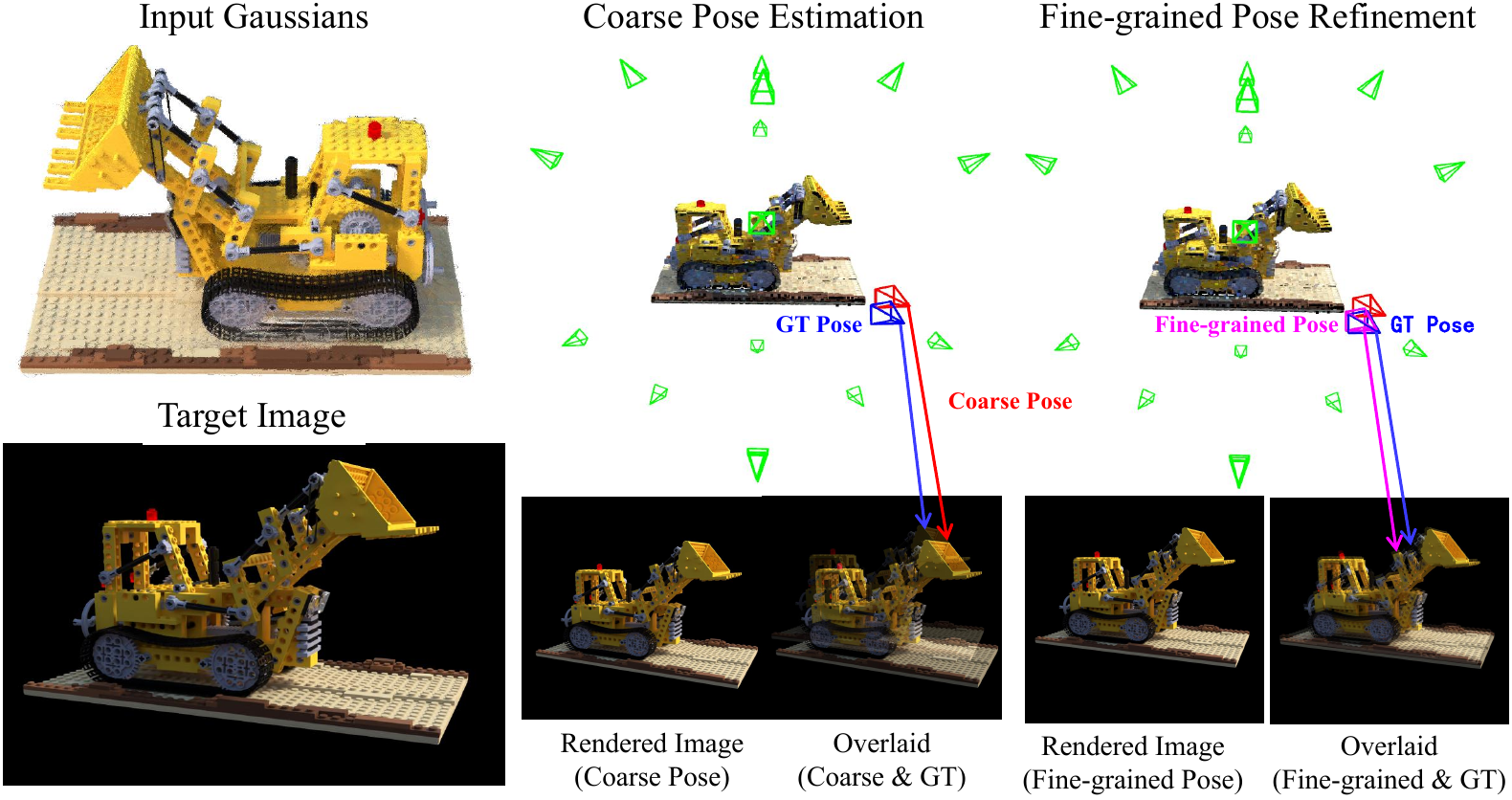}
    \vspace{-3mm}
    \caption{Results of our tow-stage pose estimation strategy. The first stage employs Gaussian Rendering-Based Coarse-Grained Pose Estimation to estimate coarse camera poses, while the second stage applies Correspondence-Based Pose Optimization for refinement. Blue camera poses denote ground truth, with red and purple coordinates representing coarse estimates and fine-grained optimizations, respectively. The comparison between rendered images (solid) and target images (ghosted) visualizes alignment accuracy.}
    \label{fig:4}
    \vspace{-5mm}
\end{figure}

\paragraph{Pose Estimation Results} Our method was evaluated in pose estimation tasks, measuring success rates with rotation errors $<5^{\circ}$ and translation errors $<5$ cm \cite{hodan2018bop}. The experimental results are shown in Table~\ref{tab:Synthetic}. In high-overlap views ($\pm[20^{\circ}, 40^{\circ}]$), it outperformed INeRF and iComMa in success rate and reduced runtime by nearly tenfold compared to iComMa. In low-overlap ($\pm[40^{\circ}, 80^{\circ}]$) and extremely low-overlap ($\pm[80^{\circ}, 180^{\circ}]$) scenarios, our method's advantages were more pronounced, maintaining stable success rates and runtime, while INeRF and iComMa's success rates dropped significantly due to their optimization-based strategies. Our method exhibited stronger robustness and efficiency. Specifically, the qualitative results are shown in Fig.~\ref{fig:4}.

%%%%%%%%%%%%%%%%%%%%%%%%%%%%%%%%%%%%%%%%%%%%%%%%%%%%%%%%%%%%%%%%%%%%%%%%%%%%%%%% table4
\begin{table}[h]
    \centering
    \label{tab:4}
    \raggedright
    \caption{\textbf{Ablation results on different components of our approach.} 
    % This Table show that WMP and ViT reduce translation/rotation errors by 30\%, while Matching + Solver is critical for low-disparity accuracy. Our full method achieves minimal errors (0.11m, 0.43°) when all components are enabled.
    }
    \label{tab:ablation}
    \vspace{2pt}
    \resizebox{\columnwidth}{!}{ 
    \begin{tabular}{ccccccccc}
        \toprule
        \multirow{2}{*}{Pose Attention} & \multirow{2}{*}{WMP} & \multirow{2}{*}{ViT} & {LoFTR} & \multicolumn{2}{c}{Translation (m)} & \multicolumn{2}{c}{Rotation ($^\circ$)} \\
        \cmidrule(lr){5-6} \cmidrule(lr){7-8}
        &  &  & + RANSAC & Avg.$\downarrow$ & Med.$\downarrow$ & Avg.$\downarrow$ & Med.$\downarrow$ \\
        \midrule
        $\checkmark$ & $\times$ & $\times$ & $\times$ & 0.38 & 0.3 & 5.18 & 4.94 \\
        $\checkmark$ & $\checkmark$ & $\times$ & $\times$ & 0.29 & 0.23 & 3.95 & 3.63 \\
        $\checkmark$ & $\checkmark$ & $\checkmark$ & $\times$ & 0.27 & 0.22 & 2.26 & 2.18 \\
        $\checkmark$ & $\checkmark$ & $\times$ & $\checkmark$ & 0.19 & 0.13 & \textbf{0.43} & \textbf{0.2} \\
        \textbf{$\checkmark$} & \textbf{$\checkmark$} & \textbf{$\checkmark$} & \textbf{$\checkmark$} & \textbf{0.11} & \textbf{0.08} & \textbf{0.43} & \textbf{0.2} \\
        \bottomrule
    \end{tabular}   
    }
    \vspace{-5pt}
\end{table}
%%%%%%%%%%%%%%%%%%%%%%%%%%%%%%%%%%%%%%%%%%%%%%%%%%%%%%%%%%%%%%%%%%%%%%%%%%%%%%%% table4

\subsection{Mip-NeRF 360 Dataset}
\label{subsec:B}
\setcounter{paragraph}{0} % 重新从 'a' 开始编号
\paragraph{Setting}We evaluated our method on nine scenes from the Mip-NeRF 360 dataset, generating 300 test images per scene (100 each for $\pm[20^{\circ}, 40^{\circ}]$ and $\pm[40^{\circ}, 80^{\circ}]$ $\pm[80^{\circ}, 180^{\circ}]$angle deviations). Using a $15^{\circ}$ rotation error threshold, we computed median and mean prediction errors to analyze performance. The Gaussian sampling strategy we adopted is set with $R = 0.1$, and our results outperform the existing SOTA. 
\paragraph{Pose Estimation Results}The experimental results in Table~\ref{tab:2} show that our method outperforms FAR \cite{rockwell2024far} method in both high-overlap and low-overlap views for pure rotation estimation. In low-overlap cases, FAR struggles with increased errors due to sparse matching points, while our method regresses a precise initial pose and optimizes it using a render-and-compare strategy, significantly reducing errors. Additionally, LoFTR \cite{sun2021loftr} excels in low-parallax views but lacks translation scale information, so we integrated 8-Point ViT \cite{rockwell20228} for scale estimation. 

\subsection{T\&T+DB Dataset}
\label{subsec:C}
\setcounter{paragraph}{0} % 重新从 'a' 开始编号
\paragraph{Setting}We evaluated our method on four scenes from the T\&T+DB dataset, generating the test dataset using the Mip-NeRF 360 strategy with Gaussian sampling ($R$=0.1) and viewpoints facing away from the origin. We investigated the performance of the model under the conditions of a 15° rotation error threshold and a 0.5m translation error threshold. We compared our method with solver-based (LoFTR, SuperGlue) and learning-based (ViT-8-Point), as well as plane mapping and optimization-based methods (NOPE-SAC \cite{tan2023nope}). Additionally, we benchmarked against FAR, which integrates optimization and learning paradigms and outperforms standalone methods, making it a key reference in our study.
\paragraph{Pose Estimation Results}Table~\ref{tab:3} shows our method's performance on the T\&T+DB dataset. FAR, combining matching and solver-based paradigms, excels in various metrics, while end-to-end methods like 8-Point ViT perform well in translation estimation, and solver-based methods like LoFTR lead in rotation estimation. Our method surpasses FAR, reducing mean and median translation errors by $\sim$50\% (0.24 to 0.11 and 0.15 to 0.08) and rotation errors by $\sim$50\% (1.23 to 0.43) in the $\pm[20^{\circ}, 40^{\circ}]$ range. In large disparity views, it outperforms FAR by over 8$\times$ in both translation and rotation, demonstrating high accuracy, robustness, and generalization under extreme viewpoint changes.

\begin{figure}[t]
    \centering
    \includegraphics[width=\linewidth]{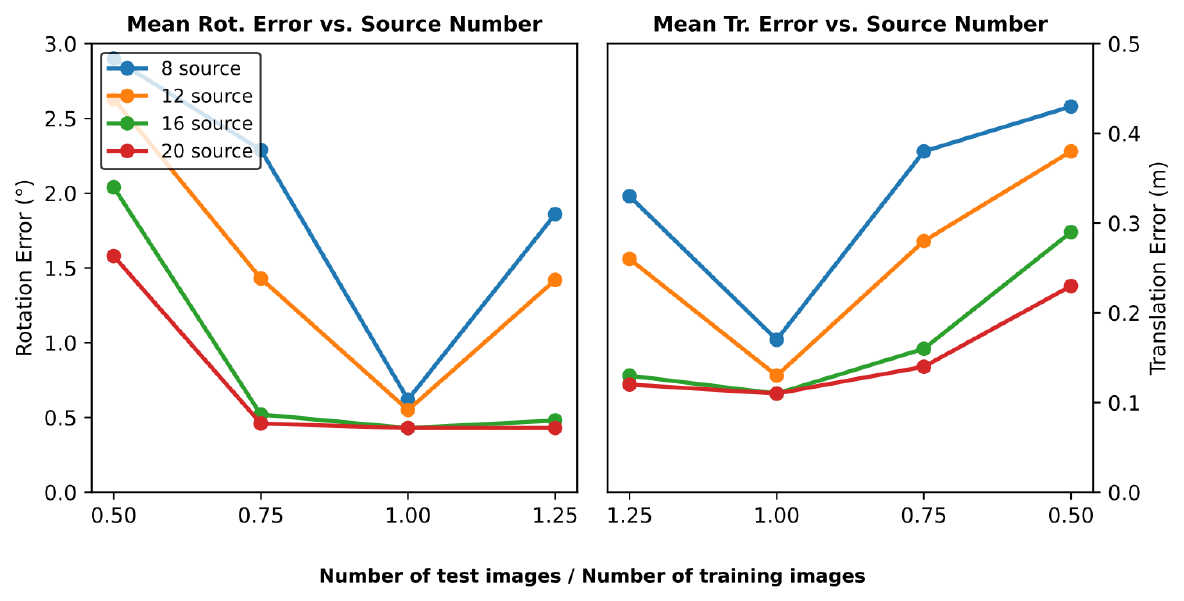}
    \vspace{-1.2em}
    \caption{Generalization Test on T\&T+DB. We evaluate the model’s performance under varying numbers of input images to determine the optimal quantity for training and assess its robustness to input variations.}
    \label{fig:5}
    \vspace{-8pt}
\end{figure}

\subsection{Ablation Study}
\label{subsec:D}
\setcounter{paragraph}{0} % 重新从 'a' 开始编号
\paragraph{Setting}
We conducted an ablation study on the T\&T+DB dataset to analyze key components of our method: Pose Attention, Weight Predictor Module(WMP), ViT \cite{rockwell20228}, and Matching + Solver (LoFTR + RANSAC). For Stage 1, we evaluated initial pose regression using Gaussian scene priors. For ViT ablation, we assessed its role in refining translation scale estimation by applying the Stage 1 (Rendering-Based Coarse-Grained Pose Estimation) pose ($T_{coarse}$) as the scale factor. For Matching + Solver, we examined its impact on final pose accuracy. In the WMP ablation, we sampled only one source image to assess its effect on matching quality. All experiments were conducted on image pairs with $\pm[20^{\circ}, 40^{\circ}]$ angular deviation for consistency.
\paragraph{Ablation Study Results}
Table~\ref{tab:ablation} demonstrates the impact of different modules on our method’s performance, $\checkmark$ indicates the module is used, and $\times$ indicates it is not. Key findings reveal that using only Pose Attention with a single source image for coarse pose regression limits accuracy due to insufficient viewpoints, while adding WMP and increasing spatial sampling to 16 source images reduces translation and rotation errors by nearly 30\%, highlighting the importance of broad spatial sampling. ViT significantly improves translation scale estimation by analyzing the relative pose between the observation image and the rendered image, providing accurate scale correction. Additionally, removing the feature matching and solver module reduces prediction accuracy by nearly 50\%, emphasizing its critical role, particularly in low-disparity views.

\paragraph{Generalization Test}As shown in Fig.~\ref{fig:5}, we evaluate our model's generalization capability under varying numbers of source images. We tested sampling strategies with 8, 12, 16 and 20 source images, adjusting polar ($\phi$) and azimuth ($\psi$) angles for each. Additionally, we tested views where input source images were 1/2, 3/4, and 3/2 of the original training setting. As the number of source images increases, the model performance improves. When there are 16 source images, the model achieves a balance between performance and resource usage. While performance fluctuates with varying input counts, the model remains robust with 16-20 images, but insufficient images cause significant degradation, emphasizing the need for adequate viewpoint coverage.

%%%%%%%%%%%%%%%%%%%%%%%%%%%%%%%%%%%%%%%%%%%%%%%%%%%%%%%%%%%%%%%%%%%%%%%%%%%%%%%%

\section{Conclusion}
In this paper, we propose iGaussian, a novel real-time camera pose estimation method based on feed-forward 3D Gaussian inversion. Our approach integrates a multi-view pose regression network with Gaussian scene priors and a matching and solver module to seamlessly connect coarse pose estimation and fine optimization in a single feed-forward pass, bypassing the computationally expensive iterative render-compare process used in traditional methods.
Experiments show that iGaussian performs excellently in rotation and translation prediction, and achieves a real-time speed of 2.87 FPS on mobile robots. iGaussian balances accuracy and efficiency, providing a solution for visual navigation, robot localization, and AR. 

\noindent \textbf{Acknowledgements}. This work was supported in part by the National Natural Science Foundation of China under Grant U22B2050 and Grant 62376032, in part by the China Postdoctoral Science Foundation under Grant 2025M771741, and in part by the Postdoctoral Fellowship Program of the China Postdoctoral Science Foundation (CPSF) under Grant GZC20251206.

\bibliographystyle{ieeetr} %ieeetr国际电气电子工程师协会期刊
\bibliography{ref} % ref就是之前建立的ref.bib文件的前缀
\addtolength{\textheight}{-12cm}   % This command serves to balance the column lengths
\end{document}